%% file: sLAW_submitted_names.tex
\documentclass[sn-basic,final=true]{sn-jnl}

\jyear{2021}%

\usepackage[T1]{fontenc} 

\theoremstyle{thmstyleone}%

\theoremstyle{thmstyletwo}%

\theoremstyle{thmstylethree}%

\usepackage{booktabs}

\hypersetup{
  final=true,
  allcolors = purple
}

\usepackage{graphicx}
\usepackage{multicol}
\definecolor{PrologPredicate}{RGB}{0,0,200}
\definecolor{PrologVar}      {RGB}{145,032,039}
\definecolor{PrologComment}  {RGB}{169,082,044}
\definecolor{PrologOther}    {rgb}{0.2,0.2,0.2}
\definecolor{PrologString}   {RGB}{070,120,200}
\usepackage{relsize}
\usepackage{listings} 
\usepackage{wrapfig}
\newcommand{\code}{\lstinline[style=MyInline]}
\lstdefinestyle{MyInline}
{
  basicstyle = \ttfamily\color{PrologOther},
  breaklines = true,
  breakatwhitespace=true,
  upquote = true,
  literate =
  {,}{}{0\discretionary{,}{}{,}}
  {\ │}{{$\mid$}}1 
  {|}{{$\mid$}}1
  {\\\{}{{\{}}1
  {\\\}}{{\}}}1
  {[}{{\small[}}1
  {]}{{\small]}}1
  {.=.}{{\#=}}3
  {.<.}{{\#<}}3
  {.>.}{{\#>}}3
  {.=<.}{{\#=<}}4
  {.>=.}{{\#>=}}4
  {<}{{<}}2
  {>}{{>}}2
  {=<}{{=<}}3
  {>=}{{>=}}3
  {\ \\=}{{\,\char"5C=\,}}2
  {\\=}{{\char"5C=}}2
  {?-}{{?-\,}}2
  {\\$}{{\$}}1
}
\lstdefinestyle{Tree}
{
  keywords = {},
  upquote = true,
  breaklines = true,
  postbreak=\mbox{\textcolor{red}{$\hookrightarrow$}\space},
  breakatwhitespace=true,  
  basicstyle = \relsize{-1}\ttfamily\color{PrologPredicate},
  basewidth = 0.4em,
  xrightmargin=-2cm,
  moredelim = {*[s][\color{black!40!PrologPredicate}]{\#pred}{.}},
  moredelim = {*[s][\color{black!40!PrologPredicate}]{\#show}{.}},
  moredelim = {*[s][\color{black!40!PrologPredicate}]{\#hide}{.}},
  moredelim = {*[s][\color{PrologVar}]{(}{)}},
  moredelim = {*[s][\color{PrologString}]{'}{'}},
 commentstyle = \mdseries\color{PrologComment},
  morecomment=[l]\%,
   literate     =
  {|}{{$\mid$}}1
  {\\$}{{\$}}1
  {\ │}{{$\mid$}}1,
}
\lstdefinestyle{MySCASP}
{
  keywords = {},
  upquote = true,
  basicstyle = \relsize{-0.5}\ttfamily\color{PrologPredicate},
  basewidth = 0.48em,
  moredelim = {**[is][\color{PrologComment}]{`}{`}},
  moredelim = {*[s][\color{black!40!PrologPredicate}]{\#pred}{.}},
  moredelim = {*[s][\color{black!40!PrologPredicate}]{\#show}{.}},
  moredelim = {*[s][\color{black!40!PrologPredicate}]{\#hide}{.}},
  moredelim = {*[s][\color{PrologVar}]{(}{)}},
  moredelim = {*[s][\color{PrologString}]{'}{'}},
  moredelim = {*[s][\color{PrologOther}]{:-}{.}},
  moredelim = {*[s][\color{red}]{/*}{*/}},
  commentstyle = \mdseries\color{PrologComment},
  morecomment=[l]\%,
  morecomment=[s]{/*}{*/},
  literate     =
  {|}{{$\mid$}}1
  {\ │}{{$\mid$}}1
  {[}{{\color{PrologOther}\small[}}1
  {]}{{\color{PrologOther}\small]}}1
  {\\$}{{\$}}1
  {&(}{{\color{PrologOther}(}}1
  {&)}{{\color{PrologOther})}}1
  {&.}{{.}}0
  {\\=}{{\char"5C=}}2
  {\\$}{{\$}}1,
}
\normalbaroutside  
\usepackage[
textwidth=0.14\textwidth, textsize=footnotesize]{todonotes}
\usepackage{xspace}
\newcommand{\ok}{{\bf+}\xspace}
\newcommand{\no}{$-$\xspace}
\newcommand{\?}{$?$\xspace}
\newcommand{\yes}{{\bf yes}\xspace}

\begin{document}

\title{Automated Legal Reasoning with Discretion to Act using s(LAW)}

\author{\hfill Joaqu\'in Arias \hspace{1cm} Mar Moreno-Rebato \hfill \ \\
\hfill  Jose~A. Rodriguez-García  \hspace{1cm} Sascha Ossowski \hfill \
}

\affil{\orgdiv{CETINIA}, \orgname{Universidad Rey Juan Carlos}, \city{Madrid}, \postcode{28933}, \country{Spain}}






\abstract{%

  { \let\thefootnote\relax \footnotetext{This work has been supported by grant VAE: TED2021-131295B-C33 funded by MCIN/AEI/ 10.13039/501100011033 and by the “European Union NextGenerationEU/PRTR”, by grant COSASS: PID2021-123673OB-C32 funded by MCIN/AEI/ 10.13039/501100011033 and by “ERDF A way of making Europe”, and by grant 2023/00004/004 s(LAW) funded by URJC.}}

  Automated legal reasoning and its application in smart contracts and
  automated decisions are increasingly attracting interest.
  In this context, ethical and legal concerns make it necessary for
  automated reasoners to \emph{justify} in human-understandable terms
  the advice given.
  Logic Programming, specially Answer Set Programming, has a rich
  semantics and has been used to very concisely express complex
  knowledge.
  However, modelling \emph{discretionality to act} and other vague
  concepts such as \emph{ambiguity} cannot be expressed in top-down
  execution models based on Prolog, and in bottom-up execution models
  based on ASP the justifications are incomplete and/or not scalable.
  We propose to use s(CASP), a top-down execution model for predicate
  ASP, to model vague concepts following a set of patterns.
  We have implemented a framework, called s(LAW), to model, reason,
  and justify the applicable legislation and validate it by
  translating (and benchmarking) a representative use case, the
  criteria for the admission of students in the ``Comunidad de
  Madrid''.
 
}


\begin{center}
This is the AOM of a contribution whose Version of Record has been published and is available at \\
\emph{Artificial Intelligence and Law (2023)} \\
\url{https://doi.org/10.1007/s10506-023-09376-5}
\end{center}

\maketitle

\section{Introduction}

The formal representation of legal texts to automatize reasoning
about them is well known in literature. For deterministic rules there
are several proposals, often based on logic programming
languages~\citep{ramakrishna2016dialogue,sergot1986british}.
This topic is recently gaining much attention thanks to the interest
in the so-called smart contracts, and to automated decisions by public
administrations~\citep{i2020fiables,cobbe2019administrative,2020huergo,suksi2021administrative,vestri2021inteligencia}.

Law uses a natural language that is characterized by its vagueness,
ambiguity, and open texture (which admits both restrictive and extensive
interpretations), i.e.,  there are areas of certainty and areas of
uncertainty. This contrasts with other scientific disciplines built on
the use of symbolic or mathematical languages.
Modeling legal rules with computer languages is, therefore, a complex
task, since indeterminate legal concepts,
discretionary powers, general principles of law, etc., need to be formally represented.
Perhaps counterintuitively, even in the case of \emph{regulated} procedures
the formal modeling of legal rules may raise multiple problems, because in most cases it will be necessary to interpret the legal rule (specifying
indeterminate legal concepts, applying general principles of law,
connecting some legal rules with others, etc.).
The modeling of legal rules in procedures in which \emph{discretionary}
powers are articulated, increases this complexity significantly.
To meet the challenge of automating administrative procedures, it is
necessary to duly represent the aforementioned concepts pertinent to legal language within a computer-interpretable formal language.

However, existing proposals usually fall short in adequately capturing the
ambiguity and/or administrative discretion present in applicable legislation. A good example is \emph{force majeure}.
Force majeure is a law term that must be understood as referring to
abnormal and unforeseeable circumstances which were outside the
control of the party by whom it is pleaded and the consequences of
which could not have been avoided in spite of the exercise of all due
care (see judgment Court of Justice of European Union, case Tomas
Vilkas, C-640/15, 25 January 2017).
Consider, for instance, the awarding school places in centers supported with public funds in the ``Comunidad de Madrid'' (CM), in Spain. In the corresponding legal procedure, the proximity of a school to a family's home or work address plays an
important role. This proximity is determined based on existing
educational districts, except in cases of force majeure, but these
cases are not defined a priory.
Educational districts which in the current regulations of the
Community of Madrid is the municipality as a whole or each of the city
districts in the municipality of Madrid.

In this work we present a framework, called s(LAW), that allows for
modeling legal rules involving ambiguity, and supports reasoning and 
inferring conclusions based on them.
Additionally, thanks to the goal-directed execution of s(CASP), the
underlying system used to implement our proposal, s(LAW) provides
justification of the resulting conclusions (in natural language).

To evaluate the expressiveness of our proposal we have represented
the criteria for the admission of students in the
``Comunidad de Madrid'' in s(LAW). In particular we focus on
the procedure for awarding school places for the ``Educaci\'on  Secundaria Obligatoria'' (ESO) of centers supported with public
  funds in the CM.
The Spanish Organic Law on Education\footnote{Organic Law 2/2006, May
  3, last modified by Organic Law 3/2020, December 29} regulates, in
article 84, the criteria for the admission of students in public
centers and private subsidized centers and, in its second paragraph of
this article 84, indicates adjudication criteria.
However, since Spain is a politically decentralized country, it is the
autonomous communities (and, therefore, their educational
administrations) that have powers to develop these aspects of basic
state legislation.
The CM, in use of its powers in educational matters, establishes the
framework and general procedure for the admission of students to
educational centers supported with public funds for the ESO.\footnote{
  Decree 29/2013, of 11 April, amended by Decree 244/2021, of 29
  December, of the ``Consejo de Gobierno'', on freedom of choice of
  school in the ``Comunidad de Madrid'' and updating the admission
  criteria and their weighting; Order 1240/2013, of 17 April, of the
  ``Departamento de Educación, Juventud y Deportes'' of the
  ``Comunidad de Madrid'', amended by Order 1534/2019, of May 17 and
  by Order 592/2022, of 18 March, of the ``Consejería de Educación e
  Investigación'' of the ``Comunidad de Madrid''; Resolution of July
  31, 2013, of the ``Dirección General para la Mejora de la Calidad de
  la Educación'' (in relation to bilingual education); and Joint
  Resolution of the ``Viceconsejería de política educativa'' and
  ``Viceconsejería organización educativa'' by which instructions are
  issued to carry out the actions prior to the process of admission of
  students in centers supported with public funds for the 2022/2023
  academic year, of November 25, 2022.

  (\url{https://www.comunidad.madrid/sites/default/files/doc/educacion/resolucion_conjunta_admision_regimen_general_2023-2024_.pdf}).}

The case presented in this paper is, therefore, real case, based on
the regulations currently in force.

The present article is structured as follows. Section 2 provides a brief description of the field of goal-directed Answer Set Programming that the s(LAW) framework sets out from. Section 3 analyses the legal basis of Discretion to Act using Spanish legislation as an example. In section 4 we show how Discretion to Act, and other relevant concepts, can be modelled withing the s(LAW) automated reasoner, using the awarding of school places for the ESO as a running example. In Section 5 we describe how s(LAW) can generate natural language \emph{explanations} for its conclusions drawn from legal reasoning with Discretion to Act. Section 6 discusses related work while in Section 7 we point to future lines of work.

\section{Goal-Directed Answer Set Programming}

Our proposal relies on Answer Set Programming
(ASP)~\citep{gelfond88:stable_models} for coding legal rules. More
specifically, we use s(CASP)~\citep{scasp-iclp2018}, a goal-directed
implementation of ASP that features predicates, constraints among
non-ground variables, and uninterpreted functions.

The top-down query-driven execution strategy of s(CASP) has three
major advantages w.r.t.\ traditional ASP system: (a) it does not
require to ground the programs; (b) its execution starts with a query
and the evaluation only explores the parts of the knowledge base
relevant to the query; and (c) s(CASP) returns partial stable models
(the relevant subsets of the ASP stable models needed to support the
query) and their corresponding justification (proof tree). Thus, our
proposal automates commonsense reasoning and is scalable whereas
ground-based ASP systems do not (Section~\ref{sec:related-work}).

Additionally, s(CASP) provides a mechanism to present justifications
in natural language using a generic translation, and the possibility
of customizing them with directives that provide explanation patterns
in natural language.  Both plain text and user-friendly, expandable
HTML can be generated.
These patterns can be used with the program text itself, thereby
making it easier for experts without a programming background to
understand both the program and the results, i.e., partial model and
justification, of its execution.

\section{Legal Basis for the Discretion to Act}
\label{sec:discretion}

This section presents the first main contribution of this work, the
analysis of the administrative and political discretion to act. In
addition, we explain the limitation for automated application of the
discretion to act and propose a framework that would facilitate the
application of discretion to act.

\subsection{Political Discretion and Administrative Discretion}

This section gives a brief explanation of when political and/or
administrative discretion to act occurs and some examples.

\subsubsection{Political Discretion}

Political discretionality occurs when, either explicitly or implicitly,
political bodies or authorities at the highest political level
(governments, ministers, mayors, etc.) or senior officials of the
Administration are granted a margin of decision making of their own,
choosing between different possible alternatives. In reality, this is
politics, the ability to choose in order to pursue the general
interest.

This type of discretion is considered a type of strong, or maximum,
discretion, inherent to the political responsibility of the person
exercising it. Examples of this type of discretion would be, for
example, regulatory norms, the implementation of a public service, the
decision to connect two cities by train, the adoption of health
measures in a pandemic (requirement or not of a passport, curfew,
restrictions on mobility or on the hotel industry, etc.).

\subsubsection{Administrative Discretion}

\paragraph{Concept}

Administrative
discretionality~\citep{garcia1962,garcia2020,sanchez2021,cosculluela2021,otenyo2016}
implies a margin of free choice attributed by a rule to a Public
Administration and, within this, to an administrative
body. Discretionality implies being able to introduce subjective
criteria of valuation, to choose between different alternatives,
provided that these are equally lawful. Discretion may refer to the
convenience of acting or not, to the manner of acting, or to the
content of the action.

Administrative discretion has been contemplated by~\cite{huber1953} as
the ``Trojan horse of the rule of law''.
Much theorizing has been done in public law and has emphasized the
need to reduce discretionality as much as possible and to fight
against the immunities of power, focusing on the scope of its judicial
control.

\paragraph{Distinction with respect to regulated acts}

An administrative act is a statement issued by a Public Administration
and subject to Administrative Law. There are administrative acts that
are issued in the exercise of regulated powers and others that are
issued in the exercise of discretionary powers.

An administrative act is regulated when the Public Administration
cannot introduce any subjective criteria of assessment, i.e., it must
simply limit itself to applying the law. Examples from the Spanish administration are three-year terms granted to a civil servant, five-year teaching periods (at the
Spanish Universities), or an urban planning license.
Let's consider the urban planning license, which is a regulated but
complex procedure. Verifying whether a license application
complies with all requirements of urban planning and other urban
planning regulations is not a simple task, as the applicable legal rules
will require a complex legal interpretation. Once the requirements
have been met, the Administration -- in this case the City Council --
must limit itself to granting the license, i.e., it cannot introduce
additional requirements or rule out projects that, for example, it
dislikes from a political point of view.

On the other hand, there are discretionary acts, i.e., acts that are
issued in the exercise of discretionary powers. In these acts, the
Public Administration does not limit itself to automatically apply
the law. When the Public Administration exercises a discretionary
power, it can choose between different alternatives, provided that
these are equally valid from the perspective of the Law. Otherwise, it
would incur in an arbitrary activity, and arbitrariness is prohibited for
public authorities and, therefore, also for the Public Administration
(see, for instance, article 9.3. of the ``Constitución Española'').

\paragraph{Technical discretion}

Technical discretion is a subtype of administrative discretion. In
this case, the possibility of choosing between different alternatives
is limited by the applicable scientific or technical knowledge.
Examples would be the academic qualification of an examination or the
declaration of ruin of a building on the basis of a technical
study. Sometimes it is attributed to collegiate bodies, as may be the
case of the evaluation commissions in a competitive process to award
teaching positions or research projects.

\subsection{Applying discretion to act}

An administrative procedure is a set of acts that prepare and enable the last act of the procedure, which is generally a
resolution, such as, the granting or refusal of an aid, a subsidy, a
research project, a town planning license, or the obligation to pay a
fine/tax.

The administrative procedure may be completely regulated, e.g., (i) an
application for admission to a public university submitted by someone
who meets all requirements and provides the required
documentation, or (ii) the granting of an urban planning license, as
explained above (although it may seem a simple task, due to the
interpretation of legal norms it generally is not).

On the other hand, even for an administrative procedure in which
discretionary powers are exercised (those in which the Public
Administration can choose between different alternatives, as long as
they are equally valid for the Law), discretion is never absolute but subject to legal limits.
In every administrative procedure there is always a part (and it is a
qualitatively important part) that regulates the action of the Public
Administration. Among others, the legal rule always indicates the
administrative body that has to exercise that power, the procedure to
be followed to adopt the decision, the purpose pursued with the same
and the requirements demanded and the verification of the
``determining facts'' on which the decision is based before exercising
the discretion.

As a consequence, an administrative procedure is either entirely
regulated, or there will be a regulated part with a small
discretionary niche. For instance, in a hypothetical administrative
procedure for awarding places in public residences for the elderly, 90
percent of the points to be awarded to the different candidates would
be regulated and determined by the income level of the candidates and
their socio-health conditions, but remaining 10 percent of points
might be awarded by a multidisciplinary team based on criteria that
its members would agree upon.  Another example is the awarding of
school places discussed in the paper. It refers to regulated
requirements, with the exception of the small discretionary score
that can be awarded by the centers.

\subsection{Judicial control/review}

The most interesting aspect of the discretion to act is its judicial
control (\cite{sourdin2021}).

As we have explained above, discretion means choosing between
different alternatives (A, B, C, D...) as long as they are equally
valid for the law; that is, as long as they are not arbitrary,
discriminatory or illegal because they go against legal norms.

For example, in the context of our running use case, procedure for
awarding school places in publicly funded centers:

\begin{itemize}
\item   Schools may award points for having siblings in the school, or
  because the parents have been former students, or because the mother
  has been a victim of gender-based violence and resides near the
  school,

\item   but not for belonging to one religion or another, for being of a
  certain gender, or for having a high IQ.
\end{itemize}

This configuration of discretion determines that judges cannot enter
into judgment on the merits of the discretionary decision. They may
supervise the regulated part of the procedure and other elements
described below:

\begin{itemize}

\item may audit the body exercising jurisdiction,
 
\item whether the procedure has been followed,
 
\item whether the public purpose pursued by the rule has been
  fulfilled,
 
\item whether the indeterminate legal concepts (those that cannot be
  configured a priori and that, case by case, must be analyzed whether
  or not they are present, such as ``force majeure'') have been
  correctly applied,
 
\item whether the general principles of law have been respected, such
  as equality, legal certainty, proportionality...,
 
\item of course, also whether the requirements and the ``determining
  facts'' have been correctly applied: Did the family really live near
  the school? Was it a large family or a single parent family?

\end{itemize}

However, once all this has been checked, the judge will not be able to
assess why, out of all the options, A, B, C, D... has been chosen.
This is so because, as we mentioned before, it is a matter of choosing
between alternatives that are equally valid for the law and,
therefore, are not subject to judicial review.

On the other hand, technical discretion is also reviewable. A judicial
review can also be undertaken if a technical error committed is proven
by evidence.

The decisions of a collegial body exercising technical discretion (a
commission composed of technical experts, for example, an evaluation
commission for the awarding of teaching positions or research
projects) are presumed to be correct, unless a serious and manifest
error (arbitrariness) is demonstrated.

Finally, judicial control of political discretion is limited, since it
is a political decision and therefore subjective. However, it will not
be exempt from legal control in terms of legal compliance,
in particular, respect for human rights.

\subsection{Automate (or not) the discretion to act}

The automation of administrative decisions, even if they are
regulated, is not an easy task because a huge part of relevant evidence is not directly deducible in the available facts (with some exceptions, e.g., when it comes to computing five years of active service to recognize a five-year teaching period in Spanish Universities). Most of the procedures require an interpretation of the legal rules in order to be
applied, even if only regulated powers are exercised.  In Spain there is a doctrinal majority (for the moment) that does not
admit the use of fully automated decisions (without human
intervention) when discretionary powers are exercised. Other legal
systems that do not admit it at
all~\citep{moreno2021,perry2017,huggins2021}.

By consequence, there will be administrative procedures that can be 100\% automated
and others that can be partially automated and open a range of options
(discretionary aspects).

In this paper we propose a framework that automates the verification
of the regulated issues but also offers (duly justified) the different
legal options among which the competent body can apply discretion.

In particular, with regard to the example of the allocation of school
places discussed in the next Section, the discretionary part is
attributed to the educational centers, which are administrative
management bodies without political relevance.

\section{s(LAW): A Legal Reasoner}

In this section we present the other two contributions of the work:
(i) a set of patterns to translate legal rules into ASP, and to
generate readable justifications in natural language, and (ii) a
framework for modeling, reasoning, and justifying conclusions based on
the evidence provided by the user and the applicable law, representing
discretion, ambiguity and/or incomplete information (key concepts in
legal cases).

\subsection{Running example}

Let us consider the following use case to explain the set of patterns
and the framework we are proposing.

As we mentioned before,
the first case to be modeled in this paper refers to the allocation of
school places. Although there is freedom of choice of educational
center, when the number of applications for a given center is greater
than the number of places, the regulations establish ordering criteria
to assign points, such as the priority and complementary criteria:

\begin{itemize}
\item 
  Among the \emph{priority criteria} are the fact of having siblings enrolled
  in the same center, the proximity of the family home or place of
  work of the parents or legal guardians of the student, and the
  income of the family unit (if they are beneficiaries of a minimum
  living income or a minimum insertion income).
  
\item 
  On the other hand, \emph{complementary criteria} comprise facts like belonging to
  a large or single-parent family, parents working at the center, the
  existence of a disability of the student, being a victim of gender
  violence or terrorism, and parents who have been former students of
  the center.
\end{itemize}

Within these criteria, a small discretionary margin is attributed for
each center to award points for ``other circumstances'' that may be
established by each center, which may coincide with any of the above
or which may be decided at the discretion of each center. A series of
tie-breaking criteria are also specified. This information is not
known because it is not known whether or not there are ties. Once this
information is known, it is resolved by drawing lots. Finally, it is
established that in ``specific cases'' that are necessary to meet
educational needs, the number of groups and school units may be
increased, e.g., a case of ``force majeure'' (a meteorological
catastrophe that makes it impossible to open or use a certain center)
and, all this, because always and in any case, the schooling of all
students must be guaranteed (even if it is a different center than the
one initially chosen).

It is, therefore, a fairly regulated procedure, in which only a small
percentage of points are at the discretion of the school.

\begin{figure}
  \centering
  \begin{multicols}{2}
    \input{code_ESO}
  \end{multicols}
  \caption{Translation of the procedure for awarding school places under s(LAW).}
  \label{fig:absence}
\end{figure}


\subsection{Patterns to translate law into ASP}
\label{sec:patterns}

The translation of legal rules into logic predicates has been
considered a straightforward task for many years.
However, the translation of ambiguity and/or discretion concepts
required the help of an expert in law and/or in the field of
application, in order to specify only one interpretation and/or
decision.

Let us use the encoding of the procedure for the adjudication of
school places in the CM (Fig.~\ref{fig:absence}) to explain the
following patterns:

\paragraph{\bf Requirement For Applying}

These are the most common constructions in legal articles, and we
consider two main patterns:

\begin{itemize}
\item Disjunction of requirements, e.g., ``s/he obtains a school place
  if one of the following common requirements are met''. This is
  expressed by separating each requirement in different clauses, see
  Fig.~\ref{fig:absence} lines 9, 12, and 19:


\item Conjunction of requirements, e.g., ``In addition, some of the
  specific requirements must be met'', which is translated into a single
  clause where the comma \code{','} means \emph{and}, see
  Fig.~\ref{fig:absence} lines 5-7:

  
\end{itemize}

\paragraph{\bf Exceptions For Applying}

As we mentioned before, a legal article is, in general, a default rule
subject to possible exceptions.
In s(CASP) the exceptions can be encoded using negation as
failure. For example, Fig.~\ref{fig:absence} lines 2-4 shows the
translation of ``It will be possible to obtain a school place if the
requirement is met and there is no exception'' and then, the compiler
of s(CASP) would generate its dual, i.e., \code{not exception}, by
collecting and checking that no exceptions hold:

\begin{lstlisting}[style=MySCASP]
not exception :- not exception_1, $\color{black}{\dots}$, not exception_n.
\end{lstlisting}

\noindent
where \code{not exception_i} is a new predicate name that identified
the dual of the i$^{th}$ exception. For the sake of brevity let us
omit the explanation of how the compiler generates the dual for each
exception (see~\citep{marple2017computing,scasp-iclp2018} for details).
Fig.~\ref{fig:absence} lines 46-57 shows the translation of the unique
exception defined in our running example: ``Students
coming from non-bilingual public schools, who apply for a place in
English language bilingual schools and who wish to study in the
Bilingual Section, need to accredit a level of English in the
four skills equivalent to level B1 for 1$^{st}$/2$^{nd}$ ESO, and to level
B2 for 3$^{rd}$/4$^{th}$ ESO''.

%

\paragraph{\bf Ambiguity}

Ambiguity occurs when some aspects of the law can be interpreted in
different ways.
For example, ``proximity to the family or work address'' is a specific
and defined requirement based on the distribution by educational
districts. However, in case of \emph{force majeure}, students from an
education district may be reassigned to a school from another
district.
Fig.~\ref{fig:absence} lines 34-44 encode this scenario allowing
evaluation without having to determine a priori the force majeure
circumstances necessary to justify the reassignment of students.
This pattern generates a model where \code{force_majeure} is assumed
to hold and another model where there is \emph{no} evidence that
\code{force_majeure} holds.
%




\paragraph{\bf Discretion To Act}

Discretion to act introduces the possibility of choosing between
different options that we intend to model by generating multiple
models.
Implementations based on Prolog compute a single, canonical model, and
therefore, bypass this nondeterminism by selecting one
interpretation.
The discretion to act can be considered as a ground or an exception
following the previous patterns.
For example, Fig.~\ref{fig:absence} lines 59-79 shows the translation
of the discretion to act rule: ``The School Council may add another
complementary criterion''.
The resulting encoding uses predicates in which the variable \code{CC}
can be instantiated with different values. This feature allows us to
reuse some of the clauses without repeating them, i.e., the clauses in
lines 59-79 are generic, while clauses 81-88 specify the ground and
exceptions of the criteria added by a particular school.
Clauses in lines 66-71 generate two possible models if the discretion
to act is exercised according to the purpose / intention of the law
and it is not unlawful. In one model the complementary criterion is
applied and in the other it does not.
Then, clauses in lines 86-88 state the cases in which the discretion
to act has a purpose and/or is unlawful.

\paragraph{\bf Unknown Information}

The use of default negation may introduce unexpected results in the
absence of information (positive and/or negative).
Therefore, in many cases the desirable behavior should capture the
absence of information by generating different models depending on the
relevant information. For example, it may be unclear whether the
documents we have to certify that we are a \code{large_family} are
valid or not, so we avoid introducing that information and the
reasoner would reason assuming both scenarios.
To state that some information is certain we would use the predicate
\code{evidence/2}, e.g., \code{evidence(st01, large_family)} means
that student 1 has the condition of large family.
Additionally, s(LAW) would provide \emph{strong} negation, denoted
with \code{'-'}, to specify that we have evidences supporting the
falsehood of some information, e.g., \code{-evidence(st04, large_family)}
means that student 4 does not have the condition of large family.


%

\subsection{Description of s(LAW)}
\label{sec:description-slaw}

s(LAW), built on top of s(CASP), is composed by three modules: the
first contains the \emph{articles}, the second contains
\emph{explanations} to generate readable justifications, and the third
one contains \emph{evidence} from a set of students. In our running
example:

\paragraph{\bf ArticleESO.pl}
Contains the legislation rules in Fig.~\ref{fig:absence} following the
patterns described in Section~\ref{sec:patterns}.

\paragraph{\bf ArticleESO.pre.pl}
Contains the natural language patterns for the predicates that are
relevant to provide readable justifications of the conclusions
inferred by s(LAW).
The directive \code{#pred} defines the natural language patterns,
e.g.:

\begin{lstlisting}[style=MySCASP]
#pred obtain_place(St) :: '@(St) may obtain a school place'.
\end{lstlisting}

\noindent
Note that the natural language pattern is customized based on the id
of the student, \code{@(St)}.

Additionally, to facilitate the understanding of the code we can
obtain a readable code (in natural language) by invoking %
\code{scasp --code --human}.

\paragraph{\bf Students.pl}

\begin{figure}
\begin{multicols}{2}
\begin{lstlisting}[style=MySCASP,xleftmargin=.7cm,basewidth=.43em]
#include('ArticleESO.pl').
#include('ArticleESO.pred.pl').

come_non_bilingual(St).
want_bilingual_section(St,'2nd ESO').

student(st01).
evidence(st01, large_family).
evidence(st01, renta_minima_insercion).
evidence(st01, sibling_enroll_center).
evidence(st01, same_education_district).
evidence(st01, b1_certificate).
-evidence(st01, foreign_student).
-evidence(st01, specific_etnia).

student(st02).
...
\end{lstlisting}
\end{multicols}
\caption{First lines of the file \code|students.pl|.}
\label{fig:students}
\end{figure}
Fig.~\ref{fig:students} shows the encoding of the module
\code{students.pl} corresponding to a set of 6 student. This last
module encodes the evidence of 6 students and links them with the
previous modules \code{ArticleESO.pl} and \code{ArticleESO.pred.pl}
(lines 1-2).
The predicates \code{evidence/2} and \code{-evidence/2} (explained in
Section~\ref{sec:patterns}) are used to specify the known information,
second argument, (positive or negative evidences) for each student,
first argument.
For the sake of brevity, let us handle as \emph{unknown} evidence
corresponding to: \code{large_family}, \code|renta_minima_insercion|,
\code|sibling_enroll_center|, \code|same_education_district|,
\code|b1_certificate|, \code|foreign_student|, and
\code|specific_etnia|. Fig.~\ref{fig:students} lines 7-14 provide the
known information corresponding to the first student.
Note that we consider that all students, coming from non-bilingual
public schools, apply for a place in English language bilingual
schools and wish to study in the Bilingual Section
(Fig.~\ref{fig:students} lines 4-5).

\section{Explainable Reasoning}

\label{sec:real-appl-reas}

The modules of s(LAW) are implemented under s(CASP) version 0.22.12.14
(\url{https://gitlab.software.imdea.org/ciao-lang/sCASP}), that runs
under Ciao Prolog version 1.19-480. (\url{http://ciao-lang.org/}).
The benchmarks used in this section are available at  \url{http://platon.etsii.urjc.es/~jarias/papers/slaw-ailaw23}
and were run on a MacOS 13.2.1 laptop with an Apple Core M2.

\begin{table}[tb]
  \centering
  \small
  \setlength{\tabcolsep}{5pt}
  \caption{Case of different students evaluated using s(LAW).\\{\small
      Note: `\ok' is a positive evidence, `\no' is a negative evidence,
      `\?' means  unkown.}}
  \label{tab:student}
  \begin{tabular}{lcccccc}
    \toprule
                                    & st01  & st02  & st03  & st04  & st05  & st06   \\
    \midrule
    \code|large_family|             & \ok & \ok & \ok & \no & \no & \no  \\
    \code|renta_minima_insercion|   & \ok & \ok & \ok & \?  & \no & \no  \\
    \noalign{\vspace {.25cm}}
    \code|sibling_enroll_center|    & \ok & \ok & \no & \ok & \no & \no  \\
    \code|same_education_district|  & \ok & \ok & \no & \ok & \no & \no  \\
    \noalign{\vspace {.25cm}}
    \code|b1_certificate|           & \ok & \no & \ok & \?  & \no & \no  \\
    \noalign{\vspace {.25cm}}
    \code|foreign_student|          & \no & \no & \no & \no & \ok & \no  \\
    \code|specific_etnia|           & \no & \no & \no & \no & \no & \ok  \\
    \midrule                                                           
    \code|?- obtain_place(stXX)|   & \yes& no  & \yes& \yes& \yes& no   \\
    \bottomrule    
  \end{tabular}
\end{table}


{\bf A priori Deduction:}
Consider we run our reasoner s(LAW) in the interactive mode to
reason about the six students by invoking:
\begin{lstlisting}[style=MySCASP]
scasp -i --tree --human --short --pos students.pl
\end{lstlisting}

Then, we launch the queries to obtain conclusions from the reasoner.
Table~\ref{tab:student} shows the data corresponding to the candidates
and the conclusion generated by s(LAW) for the query %
\code{?-obtain_place(St)}. Students 1, 3, 4, and 5 may obtain a place
at the school\footnote{We discuss later on that under different
  assumptions students 3 and 4 do not obtain a place.} while students
2 and 6 do not:

\begin{itemize}

  \begin{figure}[t]
  \centering
    \input{just01}
    \caption{Justification in Natural Language for the query \code|?-
      obtain_place(st01)|.}
  \label{fig:just}
\end{figure}

\item Student 1: Fig.~\ref{fig:students} contains the information
  corresponding to this student. Since s/he meets common and specific
  requirements and avoids the exception (having level b1 in English),
  the evaluation returns the partial model:
  
\begin{lstlisting}[style=MySCASP, numbers=none, xleftmargin=0mm]
{ obtain_place(st01),  large_family(st01),  sibling_enroll_center(st01),
  come_non_bilingual(st01), want_bilingual_section(st01,2nd ESO),
  b1_certificate(st01) }
\end{lstlisting}

  \noindent
  and the corresponding justification shown in Fig.~\ref{fig:just}.
  
\item Student 2: meets common and specific requirements but has to be
  rejected because s/he does not accredit level b1 in English (in
  Table~\ref{tab:student} that is identified with a `\no' in the
  corresponding column/row). Therefore, for the query
  \code{obtain_place(st02)} s(LAW) returns \textbf{no model}.

  \begin{figure}[t]
  \centering
    \input{just03}
    \caption{Justification in Natural Language for the query \code|?- obtain_place(st03)|.}
  \label{fig:just03}
\end{figure}

\item Student 3: meets common requirements and avoids the exception.
  But s/he does not meet any specific requirement
  (\code{sibling_enroll_center} or
  \code{same_education_district}). Nevertheless, by assuming
  \code{force_majeure} s/he also meets a specific requirement
  \code{school_proximity}, so s(CASP) returns the partial model:

\begin{lstlisting}[style=MySCASP, numbers=none, xleftmargin=0mm]
{ obtain_place(st03),  large_family(st03),  school_proximity(st03),
  force_majeure,  come_non_bilingual(st03),
  want_bilingual_section(st03,2nd ESO),  b1_certificate(st03) }
\end{lstlisting}
  
  \noindent
  and we see in the corresponding justification (see
  Fig.~\ref{fig:just03}) that \code{school_proximity} holds by
  assuming \code{force_majeure}.
  
\item Student 4: in this use-case there is absence of information
  regarding the ``renta minima de insercion''
  (\code{renta_minima_insercion}) and the English certificate
  (\code{b1_certificate}), which are marked with \? in
  Table~\ref{tab:student}. The partial model returned assumes that the
  truth values for these pieces of information are true:

\begin{lstlisting}[style=MySCASP, numbers=none , xleftmargin=0mm]
{ obtain_place(st04), renta_minima_insercion(st04),
  sibling_enroll_center(st04), not exception(st04), come_non_bilingual(st04),
  want_bilingual_section(st04,2nd ESO), b1_certificate(st04) }
\end{lstlisting}

  \noindent
  therefore, based on that assumption the student would obtain a
  place.

  \begin{figure}[t]
  \centering
    \input{just05}
    \caption{Justification in Natural Language for the query \code|?- obtain_place(st05)|.}
  \label{fig:just05}
\end{figure}

\item Student 5: now let's consider that there is a school with a
  complementary criterion for foreign students and therefore, since
  the student is a foreigner, s/he obtains a
  place. Fig.~\ref{fig:just05} shows the justification for this
  example.
\item Student 6: now we consider that the complementary criterion is
  for student of a specific etnia, and that student 6 belongs to this
  ethnic group. However, this criterion, \code{specific_etnia}, cannot
  be applied because it discriminates by race and, thus, is
  unlawful. Therefore, \textbf{no model} is returned, and the student
  does not obtain a place.
\end{itemize}

{\bf A posteriori Deduction:}
The main advantage of s(LAW) is its ability to generate justifications
not only for positive but also for negative information. To extract a
justification including the negated literals we include the flag
\code{--neg} in the invocation:
\begin{lstlisting}[style=MySCASP]
scasp -i --tree --human --short --neg students.pl
\end{lstlisting}

This ability would allow us to analyze the reason for a specific
inference and/or to determine which are the requirements needed to
obtain a specific conclusion:

\begin{itemize}

  \begin{figure}[t]
    \input{neg_just02}
    \caption{Justification in Natural Language for the query \code|?- not obtain_place(st02)|.}
  \label{fig:neg_just02}
\end{figure}

\item For student 2, the query %
  \code{?-not obtain_place} returns a partial model and the
  justification (see Fig.~\ref{fig:neg_just02}) supporting that the
  student does not obtain a place.
  While student 2 met a common and a specific requirement, s/he does
  not accredit the required level of English and s/he does not meet
  any complementary criterion. Note that the last check is done for
  every possible complementary criterion, i.e., by checking that there
  is no evidence for \code{Var0} not equal \code{foreign_student}, nor
  \code{specific_etnia}, and by checking that there is no evidence for
  both of them.
  
\item For student 3, the query %
  \code{?-not force_majeure, $ $ obtain_place(st03)} avoids the
  assumption of force majeure and therefore, s(LAW) returns \textbf{no
    model}.
  Note that student 3 does not meet any specific requirement so while
  under the assumption of \code{force_majeure} s/he meets the specific
  requirement for \code{school_proximity}, without that assumption
  s/he does not and therefore, s/he does not obtain a place.

  \begin{figure}[t]
    \input{neg_just04}
    \caption{Justification in Natural Language for the query \code|?-
      not obtain_place(st04)|.}
  \label{fig:neg_just04}
\end{figure}

\item For student 4, the query \code{?-not obtain_place(st04)} succeeds
  considering the assumptions for which this student does not obtain a
  place. For example, Fig.~\ref{fig:neg_just04} shows the
  justification where it is assumed that \code{renta_minima_insercion}
  does not holds.

\begin{figure}[t]
  \centering
    \input{neg_just06}
    \caption{Justification in Natural Language for the query
      \code|?- not obtain_place(st06)|.}
  \label{fig:neg_just06}
\end{figure}

\item For student 6, Fig.~\ref{fig:neg_just06} shows the justification
  of the query \code{?- not obtain_place(st06)} so we can analyze more
  in detail why this student is rejected.
  While the complementary criteria for student 5
  (\code{foreign_student}) is similar to \code{specific_etnia}, the
  justification tree shows that student 6 does not obtain a place
  because the complementary criterion \code{specific_etnia} is illegal
  due to \code{race_discrimination}, see Fig.~\ref{fig:neg_just06}
  lines 16-17.

\end{itemize}

Additionally, we can collect the partial models, in which the school
place is or is not obtained, together with their justification and
analyze ``Epistemic Specifications''~\citep{gelfond1994logic}, that
is, what is true in all/some models, which partial models share
certain assumptions, etc.
This reasoning makes it possible to detect the missing information
that would change the decision from ``not obtained'' (or ``obtained''
under some assumptions) to ``obtained''.
Note that, by introducing the new evidence, the resulting
justification of s(LAW) provides an explanation in which these
evidence are used to support the decision.


\section{Related Work}
\label{sec:related-work}

In attempting to model how legal discretion is exercised,
\cite{schild2005taxonomy} and \cite{kannai2007modeling} discuss the
notion of open texture, the notion of delimitation of legal domains,
and whether decisions made in a given domain are binary.
They propose to model discretionary decision-making using three
independent axes: bounded (B) and unbounded (U), defined (D) and
undefined (U), and binary (B) and continuous (C).
The resulting classification makes it possible to select the adequate
inferencing techniques for each octant:

\begin{enumerate}
\item \textbf{Bounded, Defined, Binary (BDB)}
  All issues and rules are known, and the decision is binary. A
  rule-based approach can be use, e.g., to model the domain of driving
  offences where drivers can lose their licence by being drunk based
  on the blood alcohol level.

\item \textbf{Bounded, Defined, Continuous (BDC)}
  All issues and rules are known, but the decision is continuous. In
  these cases, it is necessary to define systems, such as the US
  Sentencing Guidelines, to alleviate sentencing disparities.

\item \textbf{Bounded, Undefined, Binary (BUB)}
  The issues are known and the decision is binary, but no knowledge is
  available about how they should be combined. Knowledge discovery,
  based on previous landmark cases, can be automated, However, machine
  learning techniques require such a large number of cases that it is
  rarely available in any legal domain.

\item \textbf{Bounded, Undefined, Continuous (BUC)}
  The task of distributing property following divorce is, in general,
  an example in which the relative importance of the different factors
  is not specified, and crucial terms are not defined. In this
  scenario of knowledge discovery, to alleviate disparities,
  continuous values are involved.
  
\item \textbf{Unbounded, Defined, Binary (UDB)}
  Only some issues and the way they combine are known and the decision
  is binary, e.g., determine if a driver is guilty under the
  drink-driving regulations based on the notion of dangerous driving.

\item \textbf{Unbounded, Defined, Continuous (UDC)}
  Determine the sentence of a guilty driver under the drink-driving
  regulations based on the notion of dangerous driving.

\item \textbf{Unbounded, Undefined, Binary (UUB)}
  Legal decision-makers in these domains exercise a great degree of
  discretion with a binary output, therefore, each decision have a
  great impact, e.g., a refugee review decision has impact on both the
  applicant and relevant family.

\item \textbf{Unbounded, Undefined, Continuous (UUC)}
  Deciding on the residence of the children of the marriage after
  divorce, in a legislation that tries to move away from the ``all or
  nothing'', mitigates the risk of unfairness but requires a greater
  effort of consistency with previous judgments.

\end{enumerate}

Tasks in the octants 1 and 2 can be modeled using proposals based on
logic programming languages such as those by
\cite{ramakrishna2016dialogue} and \cite{sergot1986british} using
deterministic rules.
To model unbounded and/or undefined tasks, in octants 3-8,
non-monotonic reasoning is needed, so we propose the use of Answer Set
Programming (ASP), a successful paradigm for developing intelligent
applications and has attracted much attention due to its
expressiveness, ability to represent knowledge, incorporate
non-monotonicity, and model combinatorial problems.
%
%
Most ASP systems follow bottom-up executions that require a grounding
phase where the variables of the program are replaced with their
possible values.
During the grounding phase, links between variables are lost and
therefore an explanation framework for these systems must face many
challenges to provide a concise justification of why a specific answer
set satisfies the rules (and which rules). The most relevant
approaches are:
off-line and on-line justifications~\citep{DBLP:journals/tplp/PontelliSE09};
Causal Graph Justification~\citep{DBLP:journals/tplp/CabalarFF14}; and
Labeled ABA-Based Answer Set Justification (LABAS)~\citep{DBLP:journals/tplp/0001T16}.
  %
%
However, these approaches are applied to grounded versions of the
programs, i.e., non-ground programs have to be grounded, and they may
produce unwieldy justifications when the non-ground program has
uninterpreted functions, consults large databases and/or requires the
representation of dense domains~\citep{scasp-iclp2018}.

Note that while in the proposal presented in 2020~\citep{blind} the
implementation used a different file for each student, in the
implementation presented in this paper (see
Section~\ref{sec:description-slaw}) the information for all students
is included in a unique file/database, \code{Students.pl}.


On the other hand, systems that follow a top-down execution can trace
which rules have been used to obtain the answers more easily.  One
such system is ErgoAI (\url{https://coherentknowledge.com}), based on
XSB~\citep{xsb-journal-2012}, that generates justification trees for
programs with variables. ErgoAI has been applied to analyze streams of
financial regulatory and policy compliance in near real-time providing
explanations in English that are fully detailed and interactively
navigable. However, default negation in ErgoAI is based on the
well-founded semantics~\citep{VanGelder91} and therefore ErgoAI is not
a framework that allows the representation of ambiguity and/or
discretion.

Finally, we would like to emphasize that explainable AI techniques for
black-box AI tools, most of them based on machine learning, are not
able to explain how variation in the input data changes the resulting
decision~\citep{2017explainable}.

\section{Conclusions}

The present work is an example for the fruitful collaboration of computer
scientists and lawyers in the context of legal knowledge engineering~\citep{susskind2017}.
In principle, there are two major ways of translating,
modeling and applying legal rules by converting them into computer
language.
On the one hand, it may be the law that is adjusted to computer languages,
in which case the role of the jurist would be to reformulate the legal
rules, trying to reduce indeterminate legal concepts, discretionality
and other ambiguous elements, approaching what is called computational
law~\citep{sergot1986british,genesereth2015,ramakrishna2016dialogue,liebwald2015,branting2017}.
An example is the translation of Rule 34 of the Legal Profession
(Professional Conduct) Rules 2015 of Singapore, which proceeds to
amend the word ``business'' in the statutory
text.\footnote{\url{https://github.com/smucclaw/r34_sCASP}.}

On the other hand, the computer languages may be capable of modeling the
legal language including ambiguities, its indeterminate legal concepts
and discretionality. With the present work we intended to advance in this second
possibility through discretionary decision support systems.\footnote{In the near future, it may become necessary advance in the direction of the first option as well. In this sense, we note the need to offer mixed curricula, Law and Artificial Intelligence~\citep{rodriguez2018}.}

In particular, in this paper we have shown that using goal-directed answer set
programming, s(LAW) is capable of modeling vague concepts such as
discretion and ambiguity:
  The adjudication of school places is a task in the third octant
  (BUB). It has a bounded domain with a certain degree of vagueness
  (it is not 100\% defined) and is a binary decision making.
%
The deduction based on s(LAW) allows: the consideration of
different conclusions (multiple models) which can be analyzed by
humans thanks to the justification generated in natural language; and
the reasoning about the set of these conclusions/models.
To the best of our knowledge, s(LAW) is the only system that exhibits
the property of modelling vague concepts.\footnote{On January
  14$^{th}$, 2021, Dr. Robert Kowalski explained how they bypassed
  in~\citep{sergot1986british} the representation of vague concepts
  such as \emph{without undue delay}~\citep[1:20:15,
  1:26:00]{KowalskiTalk}.}

Our future work unfolds among two major lines.  First, complete the
modeling of the legislation by tabulation for each of the criteria
used in the procedure for adjudication of school places in centers
supported with public funds. The use of this tabulation of criteria to
exploit the underlying constraint solver of s(CASP), and check
whether automated decisions can be made when from a continuous axes
(the tabulation criteria) the decision-making is binary.
Second, we want to explore the combination of different rule-based
techniques to create a \emph{hybrid} model where (i) the legislation
is modeled as rules, (ii) previous sentences are stored in databases
(non-monotonic reasoning is needed to ensure consistency), and (iii)
apply inductive logic programming (instead of machine learning) to
``learn'' rules from reduced number of landmark cases.

\bibliography{clip,general,ia}

\end{document}

%% file: code_ESO.tex
\begin{lstlisting}[style=MySCASP, basewidth=.45em]
%% Obtain a school place if...
obtain_place(St) :-
  met_requirement(St),
  not exception(St).
met_requirement(St) :-
  met_common_requirement(St),
  met_specific_requirement(St).
%% Common requirements:
met_common_requirement(St) :-
  large_family(St).

met_common_requirement(St) :-
  recipient_social_benefits(St).
recipient_social_benefits(St) :-
  renta_minima_insercion(St).
recipient_social_benefits(St) :-
  ingreso_minimo_vital(St).

met_common_requirement(St) :-
  disability_status(St).
disability_status(St) :-
  disabled_parent(St).
disability_status(St) :-
  disabled_sibling(St).
%%  Specific requirements:
met_specific_requirement(St) :-
  sibling_enroll_center(St).
met_specific_requirement(St) :-
  legal_guardian_work_center(St).

met_specific_requirement(St) :-
  relative_former_student(St).

met_specific_requirement(St) :-
  school_proximity(St). 
school_proximity(St) :-
  same_education_district(St).
school_proximity(St) :-
  not same_education_district(St),
  force_majeure.    % Ambiguity
force_majeure :-
  not n_force_majeure.
n_force_majeure :-
  not force_majeure.
%% Exceptions:
exception(St) :-
  come_non_bilingual(St),
  want_bilingual_section(St,Course),
  not accredit_english(St,Course).
accredit_english(St,'1st ESO') :-
  b1_certificate(St).
accredit_english(St,'2nd ESO') :-
  b1_certificate(St).
accredit_english(St,'3rd ESO') :-
  b2_certificate(St).
accredit_english(St,'4th ESO') :-
  b2_certificate(St).
%% Discretion To Act:
obtain_place(St) :-
  not met_requirement(St),
  met_complement_criterion(St,CC).
obtain_place(St) :-
  met_requirement(St), exception(St),
  met_complement_criterion(St,CC).

met_complement_criterion(St,CC) :-
  school_criteria(St,CC),
  purpose(CC), not unlawful(CC),
  not n_met_complement_criterion(St,CC).
n_met_complement_criterion(St,CC) :-
  not met_complement_criterion(St,CC).
purpose(CC) :-
  promote_diversity(CC).
unlawful(CC) :-
  sex_discrimination(CC).
unlawful(CC) :-
  race_discrimination(CC).
unlawful(CC) :-
  religion_discrimination(CC).

promote_diversity(foreign_student).
promote_diversity(specific_etnia).
race_discrimination(specific_etnia).

school_criteria(St,foreign_student) :-
  foreign_student(St).
school_criteria(St,specific_etnia) :-
  specific_etnia(St).
\end{lstlisting}


%% file: just01.tex
\begin{lstlisting}[style=Tree]
st01 may obtain a school place, because
    a common requirement is met, because
        st01 is part of a large family.
    a specific requirement is met, because
        st01 has siblings enrolled in the center.
    there is no evidence that an exception applies, because
        st01 came from a non-bilingual public school, and
        st01 came from a non-bilingual public school, justified above, and
        st01 wish to study 2nd ESO in the Bilingual Section, and
        st01 accredit required level of English for 2nd ESO, because
            in the four skills certificate level b1.
\end{lstlisting}


%% file: just03.tex
\begin{lstlisting}[style=Tree]
st03 may obtain a school place, because
    a common requirement is met, because
        st03 is part of a large family.
    a specific requirement is met, because
        the school is near the family or work, because
            'force_majeure' holds, because
                it is assumed that 'force_majeure' holds.
    there is no evidence that an exception applies, because
        st03 came from a non-bilingual public school, and
        st03 came from a non-bilingual public school, justified above, and
        st03 wish to study 2nd ESO in the Bilingual Section, and
        st03 accredit required level of English for 2nd ESO, because
            in the four skills certificate level b1.
\end{lstlisting}


%% file: just05.tex
\begin{lstlisting}[style=Tree]
st05 may obtain a school place, because
    the criterion foreign_student is met, because
        st05 meets the criteria foreign_student, because
            st05 is a foreign student.
        foreign_student follows the purpose of the procedure, because
            foreign_student promotes the diversity.
        it is assumed that the criterion foreign_student is met.
\end{lstlisting}


%% file: neg_just02.tex
\begin{lstlisting}[style=Tree]
there is no evidence that st02 may obtain a school place, because
    a common requirement is met, because
        st02 is part of a large family.
    a specific requirement is met, because
        st02 has siblings enrolled in the center.
    an exception applies, because
        st02 came from a non-bilingual public school, and
        st02 wish to study 2nd ESO in the Bilingual Section, and
        there is no evidence that st02 accredit required level of English for 2nd ESO, because
            there is no evidence that in the four skills certificate level b1.
    an exception applies, justified above, and
    there is no evidence that the complementary criterion Var0 not equal foreign_student, nor specific_etnia is met, because
        there is no evidence that st02 meets the criteria Var0 not equal foreign_$[\dots]$
    there is no evidence that the complementary criterion foreign_student is met, because
        there is no evidence that st02 meets the criteria foreign_student, because
            there is no evidence that st02 is a foreign student.
    there is no evidence that the complementary criterion specific_etnia is met, because
        there is no evidence that st02 meets the criteria specific_etnia, because
            there is no evidence that st02 belongs to a specific etnia.
\end{lstlisting}


%% file: neg_just04.tex
\begin{lstlisting}[style=Tree]
there is no evidence that st04 may obtain a school place, because
    there is no evidence that a common requirement is met, because
        there is no evidence that st04 is part of a large family, and
        there is no evidence that st04 is a recipient of the RMI, because
            it is assumed that there is no evidence that st04 is a recipient of the RMI.
        there is no evidence that a parent or sibling of st04 has disability status.
    there is no evidence that the complementary criterion Var0 not equal foreign_student, nor specific_etnia is met, because
        there is no evidence that st04 meets the criteria Var0 not equal foreign_$[\dots]$
    there is no evidence that the complementary criterion foreign_student is met, because
        there is no evidence that st04 meets the criteria foreign_student, because
            there is no evidence that st04 is a foreign student.
    there is no evidence that the complementary criterion specific_etnia is met, because
        there is no evidence that st04 meets the criteria specific_etnia, because
            there is no evidence that st04 belongs to a specific etnia.
\end{lstlisting}


%% file: neg_just06.tex
\begin{lstlisting}[style=Tree]
there is no evidence that st06 may obtain a school place, because
    there is no evidence that a common requirement is met, because
        there is no evidence that st06 is part of a large family, and
        there is no evidence that st06 is a recipient of the RMI, and
        there is no evidence that a parent or sibling of st06 has disability status.
    there is no evidence that the complementary criterion Var0 not equal foreign_student, nor specific_etnia is met, because
        there is no evidence that st06 meets the criteria Var0 not equal foreign_$[\dots]$
    there is no evidence that the complementary criterion foreign_student is met, because
        there is no evidence that st06 meets the criteria foreign_student, because
            there is no evidence that st06 is a foreign student.
    there is no evidence that the complementary criterion specific_etnia is met, because
        st06 meets the criteria specific_etnia, because
            st06 belongs to a specific etnia.
        specific_etnia follows the purpose of the procedure, because
            specific_etnia promotes the diversity.
        specific_etnia is illegal, because
            specific_etnia discriminates based on race.
\end{lstlisting}
